\documentclass{article}
\usepackage{spconf,amsmath,amssymb,graphicx,booktabs,tikz,pgfplots,hyperref}
\usetikzlibrary{positioning}
\pgfplotsset{compat=1.16}
 
\usepackage[font=small]{caption}
 
\newcommand{\pmix}{p^{\mathrm{mix}}}
\newcommand{\pbase}{p^{0}}
\newcommand{\pexp}{p^{1}}

\title{Residual Expert Routing for Text-Only Domain Adaptation \\ of Factorized Transducer ASR}
\name{\begin{tabular}{@{}c@{}}Shashi Kumar$^{1,2}$, Esaú Villatoro-Tello$^{1}$, Thibault Bañeras-Roux$^{1}$, Sergio Burdisso$^{1}$ \\Petr Motlicek$^{1,3}$,
      Kadri Hacioglu$^{4}$, Andreas Stolcke$^{4}$\end{tabular}}

\address{$^{1}$Idiap Research Institute, Martigny, Switzerland;
         $^{2}$EPFL, Lausanne, Switzerland \\
         $^{3}$Brno University of Technology, Czech Republic;
         $^{4}$Uniphore, U.S.A.}
\begin{document}
\ninept
\maketitle
\begin{abstract}
Fine-tuning the non-blank predictor of a factorized transducer on target-domain text improves recognition on that domain but degrades performance on the source domain.
A single interpolation weight between base and adapted predictors cannot resolve this: the adapted predictor helps at some prefixes and hurts at others, often within one utterance.
We propose conservative residual routing: the adapted predictor becomes a residual correction to the base, and a 16k-parameter gate decides at each prefix how much of it to admit. Closing the gate reproduces the base exactly.
On a 40-hour conversational corpus, routing retains essentially all of the text-only gain on two held-out domains (19.89 to 16.74, 20.91 to 16.21 WER) while leaving source-domain WER unchanged (12.90 to 12.93); the adapted predictor alone costs 1.08 WER.
Routing needs no per-domain tuning on target-domain audio.
Analysis reveals the gate does not behave as a simple domain-level switch: it modulates intervention through the disagreement between the two predictors, suppressing it in-domain and amplifying it out-of-domain.
\end{abstract}
\section{Introduction}
\label{sec:intro}
End-to-end models such as the neural transducer \cite{graves2012} fold acoustic,
lexical and linguistic knowledge into a single network, which removes the clean
separation between acoustic and language models that made text-only domain
adaptation cheap in hybrid systems. Much subsequent work recovers an explicit
language model (LM) at inference time: shallow fusion interpolates an
external LM into the beam search \cite{kannan2018shallow}; the density ratio
method \cite{mcdermott2019density} and internal LM estimation
\cite{meng2021ilme,meng2021ilmt} subtract an estimate of the model's implicit LM
before adding the external one; and the hybrid autoregressive transducer
\cite{variani2020hat} makes the blank/non-blank decomposition explicit, giving a well-defined internal LM score.
Recent work adapts the internal LM itself on text alone \cite{meng2022ilma} and trains it jointly with the recognizer on unpaired text \cite{meng2023jeit}.

Factorized transducers build the separation into the architecture instead. The
factorized neural transducer (FNT) \cite{chen2022fnt} splits blank and vocabulary
prediction and gives the vocabulary branch a standalone predictor that behaves as
a genuine LM and can be fine-tuned on text alone. Later work improved baseline
accuracy and the fusion of blank, acoustic and internal-LM scores
\cite{zhao2023fastfnt,guo2024ilm,liu2024ifnt}, generalized the idea to modular
blank and label decoders \cite{meng2022mhat}, and scaled the non-blank predictor
up to a pre-trained LLM \cite{deng2025transducerllama}. The recipe for a new domain is the same throughout: fine-tune the non-blank predictor on target text, then use the adapted predictor in place of the original during decoding.

The recipe works, but it trades one domain for another. In our experiments
(Sec.~\ref{sec:results}), adapting the non-blank predictor on text from two unseen
domains lowers WER on those domains by 15.1\% and 23.0\% relative, and raises WER
on the original domain by 8.4\% relative. Degradation on general test sets has
been reported since the original FNT paper
\cite{chen2022fnt,liu2024ifnt}. The usual control is a scalar interpolation weight tuned per deployment \cite{ma2023ilmeada}, but one value must serve every prefix: a text-adapted predictor is better where the target lexicon and phrasing apply and worse elsewhere, and both kinds of prefix occur inside the same utterance.

Conditional computation is the natural response, and gating a recognizer against an LM is not new: Cold Fusion \cite{sriram2018cold} trains a seq2seq decoder jointly with a fixed LM through a learned gate. Mixture-of-experts (MoE) layers \cite{jacobs1991moe,shazeer2017moe,fedus2022switch} route each input to a subset of experts, and have been applied in ASR as mixtures of speaker or domain adapters \cite{zhao2024saml}; parameter-space methods (adapters \cite{houlsby2019adapters}, LoRA \cite{hu2022lora}, task arithmetic \cite{ilharco2023task}) combine specialists offline. None fits our setting. Cold Fusion learns its gate while training the recognizer, so there is no deployed base to preserve. An MoE router selects among experts, none of which is the deployed model, so there is no exact fallback to it. Weight merging commits to one blended model before decoding begins. Adaptive fusion \cite{ma2023ilmeada} varies the weight by a fixed rule over external and internal LM scores; ours is a learned policy over the residual between two internal predictors.

This paper proposes conservative residual routing, which inverts the
framing. Rather than asking which expert should produce the output, we keep the
base non-blank predictor as the output and ask how much of the text-adapted expert's residual
correction to admit at the current prefix. Our contributions are:

\begin{itemize}\itemsep1pt
\item A learned routing policy that mixes the base and expert predictors in probability space, using a small gate that reads only the difference between their output distributions and decides, at each prefix, how much correction to admit. Clamping the gate shut exactly recovers the base model.
\item A training recipe that needs no target-domain audio: on paired source data
      the gate is trained through the full transducer objective and regularized
      to stay close to the base, and on target text alone it is supervised by the
      expert's next-token margin advantage.
\item Results on a 40-hour conversational corpus showing that routing retains
      essentially all of the text-only gain on two held-out domains while leaving
      source-domain WER unchanged, and that no static interpolation weight is
      better on all three sets at once.
\item An analysis showing that the gate does not behave as a simple
domain-level switch: trust is similar across domains when the
predictors agree, but responds differently as disagreement grows.
\end{itemize}

\section{Factorized Transducer}
\label{sec:ft}

Given acoustic features $\mathbf{x}_{1:T}$ and labels $\mathbf{y}_{1:U}$ over a
vocabulary $\mathcal{V}$, a transducer defines a distribution over alignments in a
lattice of blank and label emissions \cite{graves2012}. We follow the factorized
formulation of \cite{guo2024ilm}. An encoder produces $\mathbf{h}^{\mathrm{enc}}_t$,
and two predictors consume the label history, producing a blank state
$\mathbf{h}^{b}_u$ and a non-blank state $\mathbf{h}^{nb}_u$. The blank
probability comes from a joiner ($J$) over the encoder and blank-predictor states,
\begin{equation}
P_b(t,u)=\sigma\!\big(J(\mathbf{h}^{\mathrm{enc}}_t+\mathbf{h}^{b}_u)\big),
\end{equation}
while the non-blank distribution is the softmax of summed acoustic and
internal-LM scores,
\begin{align}
\log p_{am}(\cdot\,|\,t)   &= \log\mathrm{softmax}(W_a\mathbf{h}^{\mathrm{enc}}_t),\\
\log p_{ilm}(\cdot\,|\,y_{<u}) &= \log\mathrm{softmax}(W_l\mathbf{h}^{nb}_u),\\
P_{nb}(\cdot\,|\,t,u)      &= \mathrm{softmax}(\log p_{am}+\log p_{ilm}).
\end{align}
The non-blank predictor is therefore a standalone LM: it never sees the encoder,
and $p_{ilm}$ can be trained with a cross-entropy loss on text alone. Training
minimizes $\mathcal{L}_{rnnt}+\lambda_{ilm}\mathcal{L}_{ce}$, where
$\mathcal{L}_{ce}$ is the internal-LM cross-entropy on the paired transcripts.

At decoding, writing
$\tilde{p}=\mathrm{softmax}(\log p_{am}+\alpha\log p_{ilm})$, scores are combined
as
\begin{equation}
\begin{cases}
\log P_b & \text{blank,}\\[2pt]
\log\!\big((1-P_b)\,\tilde{p}\big)+\beta\log p_{ilm} & \text{non-blank,}
\end{cases}
\label{eq:decode}
\end{equation}
where $\alpha$ and $\beta$ are internal-LM fusion weights.

\section{Conservative Residual Routing}
\label{sec:method}

\begin{figure}[t]
\centering
\begin{tikzpicture}[
  font=\scriptsize, >=latex, line width=0.45pt,
  box/.style={draw, rounded corners=1.2pt, minimum height=5.5mm,
              align=center, inner sep=2.5pt},
  gate/.style={box, fill=black!10},
  op/.style={draw, circle, inner sep=0.4pt, minimum size=3.4mm}
]
\node[box, minimum width=15mm] (nb0) at (0.85,0.15) {Base NB\\predictor};
\node[box, minimum width=15mm] (nb1) at (3.15,0.15) {Expert NB\\predictor};
\node[box, minimum width=13mm] (enc) at (5.75,0.15) {Encoder};
\node at (2.00,-0.62) {$y_{<u}$};
\node at (5.75,-0.62) {$\mathbf{x}_t$};
\draw[->] (2.00,-0.45) -- (2.00,-0.26) -- (0.85,-0.26) -- (0.85,-0.14);
\draw[->] (2.00,-0.45) -- (2.00,-0.26) -- (3.15,-0.26) -- (3.15,-0.14);
\draw[->] (5.75,-0.45) -- (enc);
\node[gate, minimum width=20mm] (rt) at (2.00,1.25) {Residual gate $g_\phi$};
\draw[->] (1.25,0.43) -- (1.25,0.72) -- (1.60,0.72) -- (1.60,0.97);
\draw[->] (2.75,0.43) -- (2.75,0.72) -- (2.40,0.72) -- (2.40,0.97);
\node[box, minimum width=36mm] (mx) at (2.00,2.35)
     {$\pmix_u=\pbase_u+\gamma_u\,(\pexp_u-\pbase_u)$};
\draw[->] (0.45,0.43) -- (0.45,2.07);
\draw[->] (3.55,0.43) -- (3.55,2.07);
\draw[->] (2.00,1.53) -- (2.00,2.07) node[midway,right=1pt] {$\gamma_u$};
\node[box, minimum width=12mm] (pj) at (5.75,1.25) {Projection};
\node[box, minimum width=15mm] (ls) at (5.75,2.35) {LogSoftmax};
\draw[->] (enc) -- (pj);
\draw[->] (pj)  -- (ls);
\node[box, minimum width=7mm] (lg) at (2.00,3.15) {$\log$};
\node[op] (pl) at (3.95,3.15) {$+$};
\draw[->] (2.00,2.63) -- (2.00,2.92);
\draw[->] (2.42,3.15) -- (3.77,3.15);
\draw[->] (5.75,2.63) -- (5.75,3.15) -- (4.13,3.15);
\node[box, minimum width=14mm] (sm) at (3.95,3.92) {Softmax};
\draw[->] (3.95,3.30) -- (3.95,3.65);
\draw[->] (sm) -- (3.95,4.48);
\node at (3.95,4.64) {$P_{nb}$};
\node[anchor=south, inner sep=1pt] at (4.95,3.19) {$\log p_{am}$};
\end{tikzpicture}
\caption{Residual routing inside the non-blank branch of a factorized transducer. Only the shaded gate is trained; everything else is frozen. The blank branch is unchanged and omitted.}
\label{fig:arch}
\end{figure}
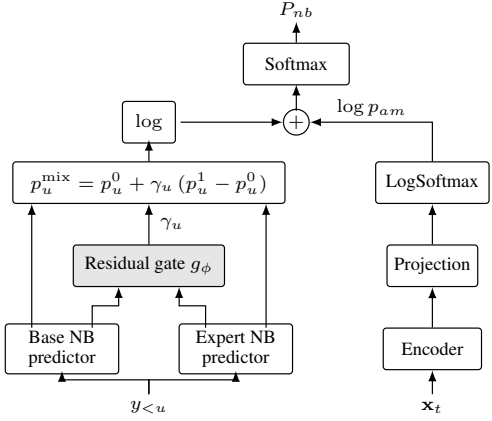

\subsection{Residual trust gate}

Let $\pbase_u(\cdot)=p_{ilm,\mathrm{base}}(\cdot\,|\,y_{<u})$ and
$\pexp_u(\cdot)=p_{ilm,\mathrm{expert}}(\cdot\,|\,y_{<u})$ be the base and
text-adapted non-blank distributions at output position $u$. The expert
contributes a residual $r_u=\pexp_u-\pbase_u$. A gate network $g_\phi$ maps that
residual to a trust value
\begin{equation}
\gamma_u=\sigma\!\big(g_\phi(r_u)\big),\qquad \gamma_u\in[0,1],
\label{eq:gate}
\end{equation}
and the mixed non-blank distribution is
\begin{equation}
\pmix_u=\pbase_u+\gamma_u r_u=\pbase_u+\gamma_u (\pexp_u-\pbase_u) .
\label{eq:mix}
\end{equation}
Because $\pbase_u$ and $\pexp_u$ are both distributions, $\pmix_u$ is a valid
distribution for any $\gamma_u\in[0,1]$, and needs
no renormalization. Its
deviation from the base is
\begin{equation}
I_u \;=\; \tfrac{1}{2}\|\pmix_u-\pbase_u\|_1 \;=\; \gamma_u \cdot D_u,
\qquad D_u=\tfrac{1}{2}\|r_u\|_1,
\label{eq:interv}
\end{equation}
so the realised intervention $I_u$ factorises into trust $\gamma_u$ and
base-expert disagreement $D_u$. This distinction matters in
Sec.~\ref{ssec:analysis}: the gate can hold trust roughly constant and still vary
its influence enormously through $D_u$.

$\pmix_u$ replaces $p_{ilm}$ everywhere in Eq.~\eqref{eq:decode}, so the gate acts within the transducer rather than as a rescoring pass (Fig.~\ref{fig:arch}).
Where the decoding rule needs log-probabilities we use
$\log(\pmix_u+\varepsilon)$. Note what the gate does not see: it has no
access to the acoustics, the identity of test sets, or either
predictor's absolute confidence. Its output where the two predictors agree
($r_u\!=\!0$) is therefore a single constant $\sigma(g_\phi(0))$ shared by every
domain, a fact we return to in Sec.~\ref{ssec:analysis}.


\subsection{Training on paired source audio-text}
\label{ssec:src}

On paired source data the gate is trained through the recognizer itself. The
mixed predictor replaces $p_{ilm}$ in the transducer, and the gate is optimized through the transducer objective with two regularizers:
\begin{equation}
\begin{aligned}
\mathcal{L}_{\mathrm{src}}={}&\lambda_{rnnt}\mathcal{L}_{rnnt}(\pmix)
 +\lambda_{ilm}\mathcal{L}_{ce}(\pmix)\\[1pt]
 &+\textstyle\sum_u\big[\lambda_{pres}\,\mathrm{KL}(\pmix_u\,\|\,\pbase_u)
 +\lambda_{sp}\,\gamma_u\big].
\end{aligned}
\label{eq:lsrc}
\end{equation}
The first two terms of Eq.~\eqref{eq:lsrc} measure how well the routed system
recognizes the source data. The transducer loss is the only term informed by the acoustics and keeps
the routed system a competent recognizer rather than a merely cautious one,
while $\mathcal{L}_{ce}$ is the internal-LM cross-entropy of Sec.~\ref{sec:ft},
now scored on the mixed predictor rather than on the base one.
The remaining two terms discourage unnecessary deviation from the base predictor. The KL anchor penalizes the realized change in the output distribution, while the sparsity term penalizes the gate value $\gamma_u$ directly.

Although the gate receives no acoustic features, source audio guides
its training through the transducer loss. For each source utterance,
this loss evaluates how well the mixed predictor and acoustic model
together explain the reference transcript. Its gradients update only
the gate; both predictors and the acoustic model remain fixed. The
gate therefore learns when admitting the expert's correction helps
recognition on source data. This complements the regularizers:
the KL and sparsity terms discourage intervention, while the
recognition loss provides a reason to intervene when the expert
is useful.

\subsection{Linguistic advantage from target text}

When target-domain text is available (with no audio), we supervise the gate with
the expert's lexical advantage. For model $i\in\{0,1\}$ define the next-token
margin
\begin{equation}
m_i(u)=\log p^{i}_u(y_u)-\max_{v\neq y_u}\log p^{i}_u(v),
\end{equation}
the linguistic advantage $A_{\mathrm{ling}}(u)=m_1(u)-m_0(u)$, and the soft target
\begin{equation}
q^{\mathrm{tgt}}_u=\sigma\!\left(\frac{A_{\mathrm{ling}}(u)}{\tau}\right).
\label{eq:bce_tgt}
\end{equation}
The gate is then trained with
\begin{equation}
\mathcal{L}_{\mathrm{tgt}}=\lambda_{ilm}\mathcal{L}_{ce}(\pmix)
 +\textstyle\sum_u\big[\lambda_{adv}\,\mathrm{BCE}(\gamma_u,q^{\mathrm{tgt}}_u)
 +\lambda_{sp}\,\gamma_u\big].
\label{eq:ltgt}
\end{equation}

This target is the natural surrogate for the decision the gate has to make.
Treating admission of the expert at position $u$ as a binary action $a$, with
conditional advantage
$\Delta(u)=\mathbb{E}[\ell(\pbase_u,y)-\ell(\pexp_u,y)\mid u]$ and preservation
cost $\mathcal{P}(u)\!\ge\!0$, the Bayes-optimal policy is the threshold rule
$a^\star(u)=\mathbb{I}\{\Delta(u)>\lambda\mathcal{P}(u)\}$: trust the expert
exactly where its advantage exceeds the cost of deviating from the base.
$A_{\mathrm{ling}}(u)$ is a differentiable proxy for $\Delta(u)$ that needs no
audio, and $\lambda_{sp}$ plays the role of $\lambda\mathcal{P}(u)$.
Using the margin rather than the reference probability matters here:
$q^{\mathrm{tgt}}_u$ is large when the expert improves the reference
token's margin over its strongest competitor relative to the base.
This captures whether the expert makes the reference more competitive,
which can help change a decoding decision for the better.
The
cross-entropy and sparsity terms appear on both sides: the mixture is always
scored against the reference tokens, and trust is only ever spent where it is
earned.

The full objective is
$\mathcal{L}=\mathcal{L}_{\mathrm{src}}+\mathcal{L}_{\mathrm{tgt}}$, with batches
drawn alternately from paired source data and target text. Only $\phi$ is
updated; the encoder, the blank predictor, both non-blank predictors, the joiner
and the output projections stay frozen.

\section{Experimental Setup}
\label{sec:setup}

\noindent\textbf{Base model.}
A Zipformer encoder \cite{yao2024zipformer} (six stacks, embedding dimensions
$192$--$512$) with a stateless blank predictor over a bigram label context and a
single-layer transformer non-blank predictor over the full label context (hidden size $512$, $8$ heads,
feed-forward dimension $2048$; $4.7$\,M parameters); $69.8$\,M parameters in
total, with a $500$-unit BPE vocabulary. It is trained with the transducer
loss \cite{graves2012} on LibriSpeech 960\,h
\cite{panayotov2015librispeech} for 20 epochs, then fine-tuned for 5 epochs on
paired audio and text from a 40-hour corpus of call-center conversations between
agents and customers (DefinedAI) \cite{kumar2026distilling,carofilis2026textonly},
covering the banking, healthcare and insurance domains.

\smallskip\noindent\textbf{Target text.}
The target data is text only, from DefinedAI \cite{carofilis2026textonly}, drawn from two domains absent from the paired
data: $10{,}981$ retail and $18{,}889$ telecommunications (Telco) utterances, $29{,}870$
in total. The same text is used for both stages below; no target-domain
audio is used to train or select any routed system.

\smallskip\noindent\textbf{Expert.}
A copy of the non-blank predictor fine-tuned on that text for 5 epochs with a
cross-entropy loss; no other module is updated.

\smallskip\noindent\textbf{Gate.}
$g_\phi$ is a two-layer neural network applied to the residual $r_u$: a linear map
from its $499$ dimensions (the BPE vocabulary without blank) to $32$
hidden units, a SiLU
nonlinearity \cite{elfwing2018sigmoid}, and a linear map to the single logit of Eq.~\eqref{eq:gate}. This
is $16{,}033$ parameters, $0.02\%$ of the full model and $0.3\%$ of the predictor
it gates. It is the only trainable module; the encoder, the blank predictor, both
non-blank predictors, the joiner and the output projections are frozen.
Training runs for 30 epochs with ScaledAdam and the Eden
schedule \cite{yao2024zipformer} at base learning rate $0.045$, alternating
source batches (up to 600\,s of audio) with target-text batches, and $\tau\!=\!1$ in Eq.~\eqref{eq:bce_tgt}. Default loss weights are
$\lambda_{rnnt}=\lambda_{ilm}=\lambda_{pres}=\lambda_{adv}=1.0$ and
$\lambda_{sp}=0.1$;
ablations vary $\lambda_{sp}$ and $\lambda_{pres}$. The checkpoint is selected by lowest
development loss on target text, which needs no target-domain audio; the
static-$\gamma$ baseline of Table~\ref{tab:main} is instead tuned by best
combined development WER, which does.

\smallskip\noindent\textbf{Evaluation.}
Three test sets: De\-fined\-AI-Retail (Retail; 3.3 h, 1312 utterances) and
De\-fined\-AI-Telco (Telco; 6 h, 2243 utterances), both out of domain with
respect to the paired data, and the in-domain test set
(DefinedAI; 3.4 h, 2009 utterances). All
systems are decoded with modified beam search at beam size $4$ and internal-LM
fusion weights $\alpha\!=\!1$, $\beta\!=\!0$ in Eq.~\eqref{eq:decode}, so that
the comparison between systems is not confounded by a second interpolation. We
report WER, and for the analysis in Sec.~\ref{ssec:analysis} the trust $\gamma_u$,
the disagreement $D_u$ and the intervention $I_u$ of Eq.~\eqref{eq:interv},
accumulated over all decoder states visited during decoding; the analysis
statistics are collected with beam size $1$ so that each utterance contributes a
single well-defined path.

\section{Results}
\label{sec:results}

\begin{table}[t]
\centering
\caption{WER (\%) on two out-of-domain (OOD) test sets and the in-domain test
set. Static interpolation holds $\gamma_u$ at a constant chosen by best combined
development WER, which needs target-domain audio; routed systems train only the
$16$k-parameter gate and are selected on target text. Best result in each column
in bold.}
\label{tab:main}
\setlength{\tabcolsep}{4pt}
\footnotesize
\begin{tabular}{@{}llccc@{}}
\toprule
 & Trained & \multicolumn{2}{c}{OOD} & In-dom. \\
\cmidrule(lr){3-4}
System & params & Retail & Telco & DefinedAI \\
\midrule
Factorized transducer (base)      & all      & 19.89 & 20.91 & 12.90 \\
\;+ text-only adaptation (expert) & pred.    & 16.88 & \textbf{16.10} & 13.98 \\
\;+ static interpolation ($\gamma\!=\!0.4$) & none & 17.16 & 17.07 & 12.89 \\
\midrule
Residual routing (default)        & gate     & 16.86 & 16.57 & 12.91 \\
\;$\lambda_{sp}=0.0$              & gate     & 16.87 & 16.53 & 12.95 \\
\;$\lambda_{sp}=1.0$              & gate     & 17.06 & 17.07 & \textbf{12.80} \\
\;$\lambda_{pres}=0.0$            & gate     & \textbf{16.74} & 16.21 & 12.93 \\
\bottomrule
\end{tabular}
\end{table}

\subsection{Baseline parity}

We first confirm that factorization does not cost performance. On LibriSpeech
test-clean/test-other, a standard transducer of matched size reaches $3.08/7.51$
WER and our factorized transducer reaches $3.05/7.40$ with $\alpha\!=\!0.7$,
$\beta\!=\!0.3$, consistent with \cite{guo2024ilm}; internal-LM fusion is what
recovers the small loss from factorization, since $\alpha\!=\!1$, $\beta\!=\!0$
gives $3.12/7.67$.

\subsection{Text-only adaptation and routing}

Table~\ref{tab:main} gives the main result. Fine-tuning the non-blank predictor on
retail and Telco text improves those domains by $15.1\%$ and
$23.0\%$ relative. It also costs $1.08$ WER in-domain, an $8.4\%$ relative
regression, the familiar price of text-only adaptation.
Residual routing keeps almost all of the gain and removes the in-domain
regression. The default configuration performs the same as the base in domain
while recovering most of the out-of-domain improvement.

Removing the KL anchor ($\lambda_{pres}\!=\!0$) improves Retail/Telco WER
from $16.86/16.57$ to $16.74/16.21$, while source-domain WER changes
only from $12.91$ to $12.93$. An explicit distribution-preservation
penalty is therefore not necessary for the observed source-domain
retention in this setting. Paired-source recognition supervision
and the sparsity penalty remain active: the former trains the gate
against recognition errors on source data, while the latter penalizes
expert admission directly. These mechanisms differ from the KL
anchor, which penalizes the resulting change in the predictor
distribution.
Increasing $\lambda_{sp}$ to $1.0$ shifts the trade-off
toward source accuracy, reaching $12.80$ WER while reducing the
target-domain gains. Conversely, removing sparsity changes WER
only slightly relative to the default, with the KL anchor and source
supervision still present. Together, these ablations suggest that
the full combination of regularizers is not required to obtain a
favorable source-target trade-off; they do not, however, isolate
the contribution of source supervision itself.


A per-prefix gate is only worth its complexity if a constant cannot do the same
job. Static interpolation, Eq.~\eqref{eq:mix} with $\gamma_u$ held at the
constant $\gamma\!=\!0.4$ selected on development data, preserves the source
domain as well as routing does but is $2.4\%$ and $5.0\%$ relative worse on
Retail and Telco. Sweeping $\gamma$ from $0.1$ to $0.9$ traces a clean trade-off in which
out-of-domain WER falls and in-domain WER rises, sharply beyond
$\gamma\!=\!0.7$, and no value is better than the routed system on all three
sets at once. The gate is buying something a scalar cannot: it adjusts the expert's contribution at each prefix, allowing
different amounts of intervention within the same utterance. The constant also
had to be chosen on target-domain audio, which the routed system never uses.

\subsection{What the gate learns}
\label{ssec:analysis}

\begin{table}[t]
\centering
\caption{Routing behaviour by test set, the first three columns restricted to
high-disagreement decoder states ($D\!\geq\!0.7$). ``Agree'' is the fraction of
those states where the
mixed and expert distributions share an argmax; ``Flip'' is the fraction of
emitted symbols differing from the base decision, at the decoder output and
after acoustic fusion and blank competition.}
\label{tab:analysis}
\setlength{\tabcolsep}{4pt}
\footnotesize
\begin{tabular}{@{}lccccc@{}}
\toprule
 & \multicolumn{3}{c}{States with $D\geq0.7$} & \multicolumn{2}{c}{Flip rate} \\
\cmidrule(lr){2-4}\cmidrule(lr){5-6}
Test set & $\mathbb{E}[\gamma]$ & $\mathbb{E}[I]$ & Agree & decoder & fused \\
\midrule
Retail (OOD)        & 0.65 & 0.55 & 0.73 & 0.234 & 0.040 \\
Telco  (OOD)        & 0.69 & 0.58 & 0.77 & 0.245 & 0.047 \\
DefinedAI (in-dom.) & 0.36 & 0.29 & 0.38 & 0.168 & 0.014 \\
\bottomrule
\end{tabular}
\end{table}

The statistics in this subsection come from the default routed system of
Table~\ref{tab:main}. The WER numbers are consistent with a gate that has learned
to recognize the target domain and switch to the expert there. The trust statistics suggest a more selective policy.
The trust distribution is strikingly similar across all three test sets: the
median $\gamma_u$ is $0.466$ on both out-of-domain sets and $0.465$ in domain,
and the distributions differ mainly in the weight of the upper tail, giving mean
trust of $0.49$, $0.50$ and $0.44$.
These similar trust distributions suggest that the gate does not
simply assign high trust to target-domain utterances and low trust
to source-domain utterances. Its behavior becomes clearer when
trust is examined together with predictor disagreement.

\begin{figure}[t]
\centering
\includegraphics[width=\linewidth]{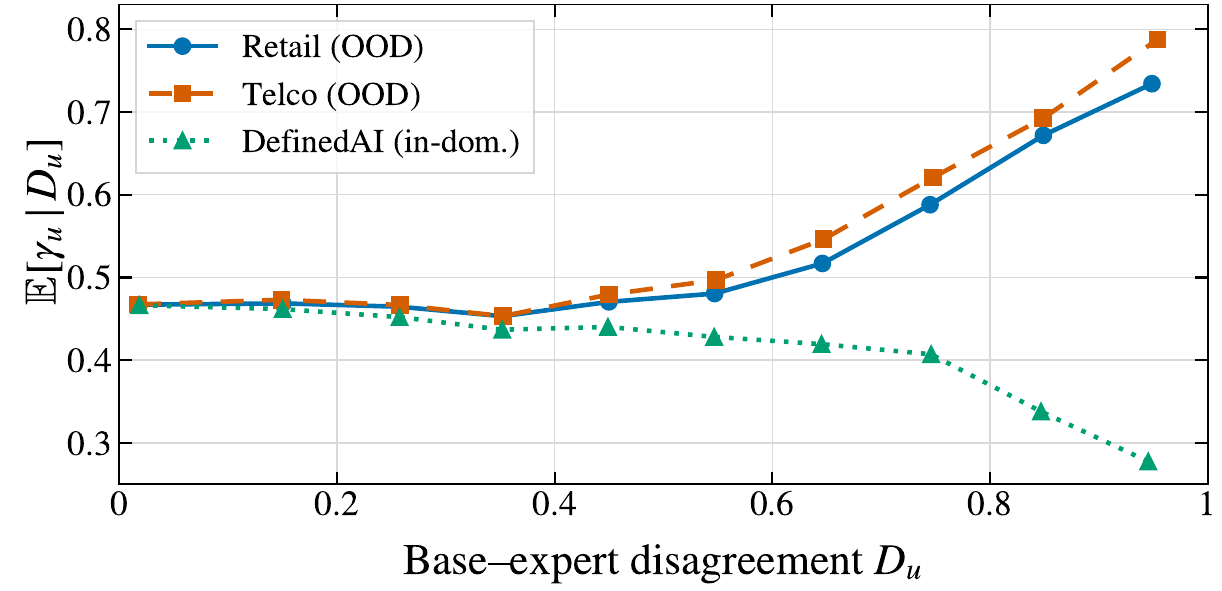}
\caption{Mean trust conditioned on base-expert disagreement, over all decoder
states visited during decoding. The curves share a starting point, since the gate
has no input to act on where the two predictors agree, and then separate: trust
grows with disagreement out of domain and shrinks with it in domain. The gate
reads the same signal as specialization on target-like data and as risk on
source-like data.}
\label{fig:trustvsd}
\end{figure}

Following
Eq.~\eqref{eq:interv}, we bin decoder states by $D_u$ and measure
$\mathbb{E}[\gamma_u\,|\,D_u]$, shown in Fig.~\ref{fig:trustvsd}. All three
curves start from the same point, as they must: where the predictors agree the
gate has no input to act on and returns $\sigma(g_\phi(0))\!=\!0.466$. They then
diverge. On both out-of-domain sets the relationship is increasing, with trust
climbing steeply at high disagreement; on the in-domain set it inverts, falling
well below its agreement value. The gate reads the same signal in
opposite directions: strong disagreement is risk on source-like data and
specialization on target-like data. Since intervention is the product
$\gamma_u D_u$, this conditional response is what separates the routed system
from any constant $\gamma$. It also rules out any policy that reads disagreement
magnitude alone: at matched $D_u$ the three curves differ, so the gate must be
responding to the direction of $r_u$ and not only its size. It is given no domain
label, and infers from the shape of the disagreement what kind of data it is on.

Table~\ref{tab:analysis} quantifies the effect in the states that matter most.
Restricted to $D\!\geq\!0.7$, mean trust is roughly twice as high out of domain as
in domain, mean intervention likewise, and the mixed distribution agrees with the
expert's top choice about twice as often. At the utterance level, most
utterances in every set fall between the two extremes, again unlike a switch.

Finally, the transducer architecture damps the gate's influence considerably.
Routing changes the decoder's top-1 choice on roughly a fifth of states, but
after the acoustic score is added and blank competition resolved, under five
percent of emitted symbols differ from the base decision
(Table~\ref{tab:analysis}). Routing edits a small, targeted subset of the output,
consistent with a $15$--$21\%$ relative WER change out of domain alongside
unchanged performance in domain.

\section{Conclusion}
\label{sec:conclusion}

We presented conservative residual routing for factorized transducers: a
text-adapted non-blank predictor becomes a residual correction to the base
predictor, and a 16k-parameter per-prefix gate decides how much of that
correction to admit. Because mixing happens in probability space, clamping the
gate shut recovers the base model exactly. On two held-out domains the routed system retains
essentially all of the text-only gain while leaving source-domain WER unchanged,
which no static interpolation weight matches on all three sets at once, and it
needs no target-domain audio to tune. The learned policy does not behave as a simple domain-level switch:
at high predictor disagreement, it assigns greater trust to the
expert on target-domain data and less on source-domain data. Nothing in
the formulation is specific to the transducer beyond the existence of a
separable output-side language model, so porting it to LLM-based transducers
\cite{deng2025transducerllama} or encoder-decoder ASR \cite{radford2023whisper}
is the natural next step.


\section{Acknowledgments}
This work was supported by Idiap Research Institute and Uniphore collaboration project. Part of this work was also supported by EU Horizon 2020 project ELOQUENCE (grant number 101070558).

\section{Compliance with Ethical Standards}
This research did not involve the collection of new data from human
participants or animals. The experiments were conducted using existing
datasets; therefore, no ethical approval was required.
\bibliographystyle{IEEEbib}
\bibliography{strings,refs}

\end{document}